\documentclass{article}
\usepackage{spconf,amsmath,graphicx}
\usepackage{booktabs} 
\usepackage{multirow}
\usepackage{caption,hyperref}

\newcommand{\etal}{\textit{et al.}}
\DeclareMathOperator*{\argmin}{argmin}

\title{Learning to compose 6-DoF omnidirectional videos using multi-sphere images}
\name{Jisheng Li$^*$,~Yuze He$^*$,~Yubin Hu$^*$,~Yuxing Han$^\dag$,~Jiangtao Wen$^*$}
\address{$^*$Tsinghua University, Beijing, China \\
        $^\dag$Research Institute of Tsinghua University in Shenzhen, Shenzhen, China}
\begin{document}
\ninept
%

\maketitle

\begin{abstract}
Omnidirectional video is an essential component of Virtual Reality. Although various methods have been proposed to generate content that can be viewed with six degrees of freedom (6-DoF), existing systems usually involve complex depth estimation, image in-painting or stitching pre-processing. In this paper, we propose a system that uses a 3D ConvNet to generate a multi-sphere images (MSI) representation that can be experienced in 6-DoF VR. The system utilizes conventional omnidirectional VR camera footage directly without the need for a depth map or segmentation mask, thereby significantly simplifying the overall complexity of the 6-DoF omnidirectional video composition. By using a newly designed weighted sphere sweep volume (WSSV) fusing technique, our approach is compatible with most panoramic VR camera setups. A ground truth generation approach for high-quality artifact-free 6-DoF contents is proposed and can be used by the research and development community for 6-DoF content generation.

\end{abstract}

\section{Introduction}
\label{sec:intro}

As a result of the rapidly increasing popularity of omnidirectional videos on video-sharing platforms such as YouTube and Veer, 
professional content producers are striving to produce better viewing experiences with higher resolution, new interaction mechanisms, as well as more degrees of viewing freedom,
using playback platforms such as the Google \textit{Welcome to Light Field}, where the viewer has 
the freedom to move along three axes with motion parallax. A body of research has been dedicated to composing 6-DoF from footage captured by VR cameras, 
extending toolsets for the professional content generators to produce more immersive contents. However, most of the proposed systems involve complex procedures such as 
depth estimation and in-painting, that are both time- and resource- consuming and require tedious hand-optimizations. 

Recently, Attal~\etal proposed MatryODShka~\cite{attal2020matryodshka}, which used a convolutional neural network to predict multi-sphere images (MSIs) that 
can be viewed in 6-DoF. This approach significantly simplified the overall pipeline complexity while producing promising visual results. 
However, the system requires omnidirectional stereo (ODS) inputs that cannot be acquired directly from widely available VR cameras, as ODS must be produced 
from raw panoramic VR footage using stitching with annoying visual artifacts. 
In addition, as in the case for any system design, 6-DoF content production also requires a large volume of high-quality content that can serve as the 
ground-truth, for both training purposes and performance evaluation. However, because 6-DoF contents are difficult to produce, there is no widely available high-quality 
6-DoF content that can be used as a ground-truth dataset for the community. 

The contribution of this paper is therefore two-fold: First, we propose a system that can produce large volumes of high-quality artifact-free 6-DoF data for training and performance evaluation
that serves as the basis for evaluation of the system proposed in this paper, as well as for future development by the community.
Second, we propose an algorithm for predicting MSI from VR camera footage directly, using a modified weighted sphere sweep volume fusing scheme 
with 3D ConvNet without explicit depth map or segmentation mask, thereby improving quality while lowering complexity.

\begin{figure}[tb] 
\centering 
 \includegraphics[width=0.75\hsize]{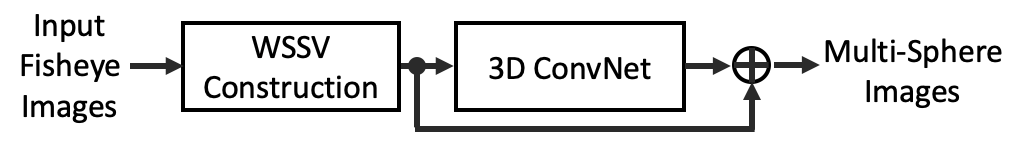}
 \caption{Our proposed 6-DoF omnidirectional video composition framework.}
 \label{fig:framework}
\end{figure}
\vspace{-9pt}
\begin{figure}[tb]
 \centering 
 \includegraphics[width=0.75\hsize]{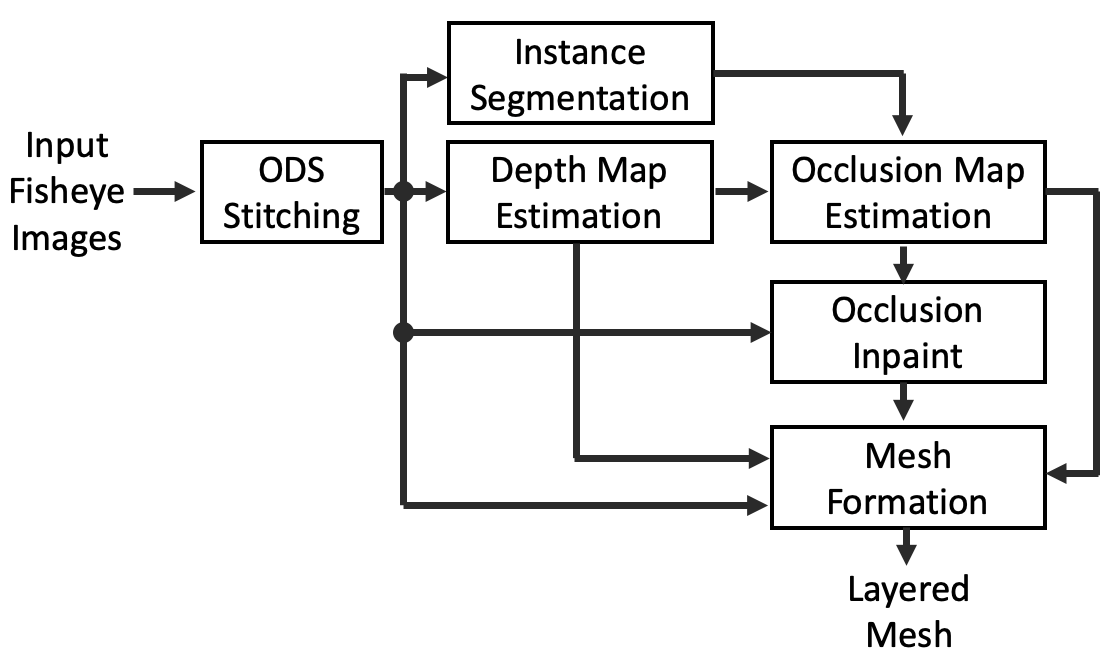}
 \caption{An example of conventional 6-DoF content generation method}
 \label{fig:conventional}
\end{figure}

\section{Related Works}
\label{sec:related_works}

Substantial research has been carried out on the reconstruction and representation of omnidirectional contents~\cite{peri1997generation}\cite{szeliski2006image}\cite{anderson2016jump}\cite{tang2018universal}\cite{bertel2020omniphotos}\cite{li2016novel}. In this section, we focus on 
recent research aimed at 6-DoF reconstruction and view synthesis. 

Six degrees of freedom content generation requires detailed scene depth information. Dynamic 3D reconstruction and content playback have 
been extensively studied in the context of free-viewpoint video, with many approaches achieving real-time performance~\cite{collet2015high}\cite{dou2016fusion4d}. 
Using the conventional multi-view stereo method, Google \textit{Welcome to Light field} ~\cite{overbeck2018welcome} and Facebook manifold system~\cite{pozo2019integrated} 
both achieved realistic high-quality 6-DoF content composition, with hardware systems that are substantially more complex than VR cameras widely available on the market. 
Lately, studies that used convolutional neural networking (CNN) show promising results for depth estimation and view synthesis~\cite{godard2017unsupervised}\cite{kalantari2016learning}\cite{srinivasan2017learning}.  
CNN achieves excellent results for predicting multi-plane images (MPI) and representing the non-Lambertian 
reflectance~\cite{zhou2018stereo}\cite{flynn2019deepview}\cite{mildenhall2019local}\cite{srinivasan2019pushing}\cite{choi2019extreme}. Many recent approaches adopt and 
extend these studies to generate omnidirectional contents~\cite{attal2020matryodshka}\cite{lin2020panorama}\cite{broxton2020immersive}. 
Broxton~\etal~\cite{broxton2020immersive} designed a half-sphere camera rig with GoPro cameras and used the method in \cite{flynn2019deepview} to generate 
pieces of MPIs and later converted it into 360 layered mesh representation for viewing. Lin~\etal~\cite{lin2020panorama} generated multiple MPIs and formed them 
into a multi-depth panorama using the MPI predicting techniques in \cite{mildenhall2019local}. Lai~\etal~\cite{lai2019real} proposed to generate panorama depth 
map for ODS images 6-DoF synthesization. More recently, MartyODShka~\cite{attal2020matryodshka} archived real time performance when
using the method in~\cite{zhou2018stereo} to synthesize MSI from ODS content. In spite of such successes, ODS images relaid by these approaches still require 
stitching of original VR camera footage, which by itself is still a problem that has not been sufficiently solved. Many existing neural network based approaches were also designed 
with a fixed number of input images that is not configurable.

In this paper, we propose an approach without stitching pre-processing or depth map estimation. In our approach, a weighted sphere sweep volume 
is fused directly from camera footage using spherical projection, eliminating artifacts introduced by stitching. By utilizing 3D ConvNet, our framework is applicable to 
different VR camera designs with different numbers of MSI layers. We also propose a high-quality 6-DoF dataset generation approach using UnrealCV~\cite{qiu2016unrealcv} 
and Facebook Replica~\cite{straub2019replica} engine, so that our, and future systems for 6-DoF content composition can be designed and evaluated  qualitatively 
and quantitatively using our data generation scheme.
\vspace{-9pt}
\section{Omnidirectional 6-DoF Data Generation}
\label{sec:data}
Commercially available VR camera models provide a range of selection for omnidirectional content acquisition. A great amount of footage from these VR 
cameras is captured by professional photographers and enthusiasts. However, it is hard to render a ground truth from these footage, as content 
captured by such cameras need to be stitched, which by itself is still a problem that has not been completely solved.  As a result, it has been extremely difficult to 
find artifact-free high-quality 360 video datasets in this literature, which is required for the continued development and evaluation of VR and/or 6-DoF content. The few 
datasets produced by companies like Google or Facebook are limited in size, content scenarios, resolutions etc., as they were constrained by the various factors 
imposed by the time and equipment used for producing such datasets. 

It is therefore highly desirable to be able to generate any number of test clips of any use cases and different resolutions and to continue this test data generation process
as new cameras with higher resolutions become available. In this study, we use two CG rendering frameworks UnrealCV~\cite{qiu2016unrealcv} and Replica~\cite{straub2019replica} 
to generate high-quality VR datasets for both training and evaluation. The UnrealCV engine can render realistic images with lighting changes and reflections on handcrafted models. 
On the other hand, Replica engine uses indoor reconstruction models, and produces  rendered images with a similar distribution to real-world textures and dynamic ranges. 
We use the combination of two datasets to mimic real-world challenges.

To compose an omnidirectional image using the above mentioned CG engines, we first generate six ${120}^{\circ}$ pin-hole images with different poses towards 
the engine's x-y-z axes and their inverse directions. The optical centers of these virtual cameras are located at the same point, to avoid any parallax between cameras. 
We then project six pin-hole images to equirectangular projection (ERP) and blend the overlapping area to avoid aliasing. To simulate camera footage from a n-sensor VR 
camera, we render ERP footage on the poses of each sensor and mask off the external details according to the lens field of view. 
In our experiments, we selected 2000 locations to generate 6-DoF contents for network training and evaluation. The generated datasets were 
split into training subsets of 1600 locations and evaluation subsets with 400 locations. The split was implemented based on virtual camera locations with 
no overlap across the training and evaluation subsets. Contents generated from UnrealCV and Replica were mixed together during training procedure 
but evaluated separately. We rendered two resolutions (640$\times$320 and 400$\times$200) on each location for training with different numbers 
of MSI layers due to GPU memory constraints. Upon that, we rendered two fields of view (${190}^{\circ}$ and ${220}^{\circ}$) to simulate fisheye images from different VR camera modules. 
\vspace{-10pt}
\section{A Simplified System for Producing 6-DoF Content from Panoramic VR Footage}
\label{sec:methodology}
Our process of turning panoramic VR camera footage into 6-DoF playable multi-sphere images involves three steps 
as shown in Fig.~\ref{fig:framework}. We first project the fisheye images onto concentric spheres using the weighted sphere sweep method and 
generate weighted sphere sweep volume (WSSV) as described in Sec.~\ref{sec:wssv}. This 4D volume is input into the 3D ConvNet detailed in Sec.~\ref{sec:network} 
to predict the $\alpha$ channel. We combine the predicted $\alpha$ channel with WSSV to form the MSI representation. Later inside the render engine, 
the MSI is calculated following Equ.~\ref{equ:6dof} to infer per-eye views in VR. 
\vspace{-9pt}
\subsection{Multi-Sphere Images}
\label{sec:msi}
\begin{figure}[tb]
 \centering 
 \includegraphics[width=0.75\linewidth]{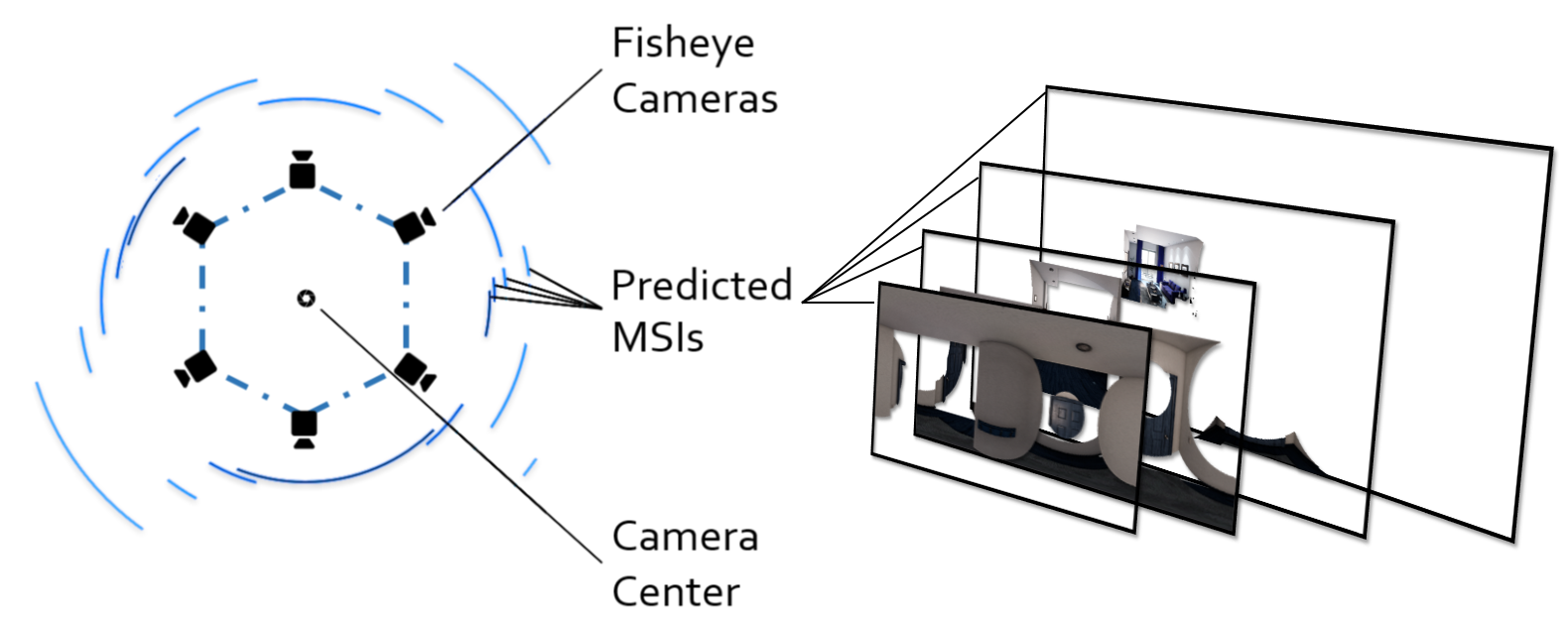}
 \caption{Illustration of an MSI representation. Blue lines show the opacity at different area of the concentric spheres.}
 \label{fig:msi}
\end{figure}
Inspired by multi-plane image (MPI) and its performance in view synthesis applications, multi-sphere images (MSIs) is generated for omnidirectional content representation. 
Following the design of MPI, the MSI representation used in our work consists of $N$ images to represent $N$ concentric spheres. These concentric spheres are warped into planes 
using equirectangular projection (ERP) for efficient processing. Each image inside an MSI contains 3 channels of color and an additional $\alpha$ channel of transparency information. 
This character inherited from MPI empowers the MSI to represent scene occlusion and non-Lambertian reflectance. The form of $RGB\alpha$ data representation 
also allows MSIs to be compressed using standard image compression algorithms.  

The differentiable rendering scheme for the MPI also applies on the MSI. As described in Equ.~\ref{equ:alpha rendering}, in MPI differentiable rendering, 
the position $\{p_i\}_{i=1}^{N}$ where a ray intersects with each layer $\{L_i\}_{i=1}^{N}$ is calculated and RGB value $\{c_{p_i}\}_{i=1}^{N}$ and  $\alpha$ value $\{\alpha_{p_i}\}_{i=1}^{N}$ 
are interpolated and then used to calculate the output color $c$
\begin{equation}
    c = \sum_{i=1}^{N}c_{p_i}\cdot{\alpha}_{p_i}\cdot \prod_{j=1}^{i-1} (1 - {\alpha}_{p_i}).
\label{equ:alpha rendering}
\end{equation}

The rendering procedure for viewing MSI in 6-DoF is similar, as described in Equ.~\ref{equ:6dof}. To render a ray inside of the inmost sphere of MSI with 
viewing angle $(\theta, \phi)$ from camera location $(x,y,z)$, the intersection point $s_{i}$ with $i$-th layers of MSI is determined by ray-plane intersection function $q(\cdot)$ in graphic engines, where ${d}_{i}$ is the radius of $i$-th sphere.
The color of the rendered ray $c_{x,y,z,\theta,\phi}$ is computed by color values $\{c_{s_{i}}\}_{i=1}^{N}$ and transparency values $\{\alpha_{s_{i}}\}_{i=1}^{N}$ in each layer following:
\begin{equation}
\label{equ:6dof}
\begin{split}
&s_{i} = q(x,y,z,\theta, \phi, d_i)\\
&{c}_{x,y,z,\theta,\phi} = \sum_{i=1}^{N}c_{s_{i}}\cdot{\alpha}_{s_{i}}\cdot \prod_{j=1}^{i-1} (1 - {\alpha}_{s_{i}}).
\end{split}
\end{equation}
This procedure can be carried out efficiently on a common CG engine such as Unity or OpenGL.

\begin{figure}[tb]
 \centering 
 \includegraphics[width=0.3\linewidth]{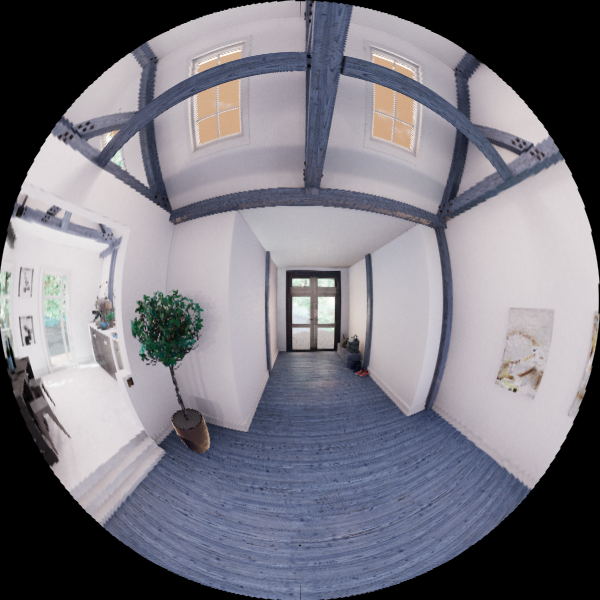}~
 \includegraphics[width=0.6\linewidth]{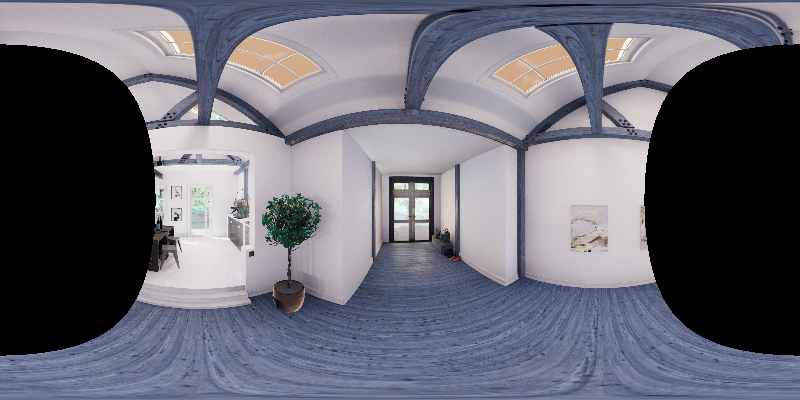}
 \caption{Left: A fisheye image with ${220}^{\circ}$ FOV. Right: A fisheye image reprojected to ERP format}
 \label{fig:fisheye}
\end{figure}
\vspace{-9pt}
\subsection{Weighted Sphere Sweep Volume Construction}
\label{sec:wssv}
A weighted sphere sweep volume is constructed with $N$ layers of concentric spheres. The radius of these spheres is chosen uniformly 
in the reciprocal space between the closest object distance and the farthest distance. As images from VR cameras are often captured using fisheye lenses that compresses 
the field of view and causes chroma aberration on the edges of images, we propose a weighted sphere sweep method to reduce such optical defects. 

To this end, we first warp these input fisheye images into the ERP form as shown in Fig.~\ref{fig:fisheye}. Then $M$ input ERP images are projected onto $N$ layers of 
sphere in ERP form using the intrinsic and extrinsic parameters of the camera. 
In each layer, we fuse $M$ input images together on the overlapping area following :
\begin{equation}
{c}_{{p}} = \sum_{i=1}^{Q}\frac{{e}^{{\gamma}_{p,i} } \cdot {c}_{p,i}}{\sum_{j=1}^{Q}{e}^{{\gamma}_{p,j}}},
\label{equ:weighted}
\end{equation}
where $Q$ is the number of overlapping images 
at position $p$, and ${c}_{p,i}$ is the color of the $i$-th image on position $p$. The parameter $\gamma$ is the optical distortion value. 
Here we use ${\gamma}_{p} = 1 - r_p$, where $r_p$ is the distance of pixel $p$ to the optical center normalized to $[0,1]$. We can also use the lens MTF data to replace $\gamma$ values
Then, $N$ projected ERP images are stacked to form a 4D volume so that the first three dimensions are ERP image height (H), width (H) and number of layers (N), while 
the 4-th dimension is the color channel. As each ERP image contains 3 channels of RGB color, the constructed WSSV has a shape of [H, W, N, 3]. 
The WSSV construction can be represented by Equ.~\ref{equ:wssv}, where the weighted and warp function ${\cal{W}}(\cdot)$ takes in a camera 
pose ${\{\omega_j\}}_{j=1}^{M}$ and the camera center pose $\omega_c$, then projects an input image from ${\{I_j\}}_{j=1}^{M}$ to a set of concentric spheres 
with predefined radius ${\{d_i\}}_{i=1}^{N}$. The ${\cal{S}}(\cdot)$ function stacks the warped results along 4-th dimension to form the weighted sphere sweep volume:
\begin{equation}
WSSV = {\cal{S}}_{j=1}^{N} ({\cal{W}}_{i=1}^{M} (I_i,\omega_i, \omega_c, {d}_{j})).
\label{equ:wssv}
\end{equation}
As each layer of sphere sweep volume is formed by multiple input images projected onto the same sphere, the combination of these images is similar to the result of light field refocusing. 
Because multiple images of the same object are projected to a layer whose radius approximately equals to the distance of that object, the average color of such a combination of the 
projected images will appear in-focus. 
The 3D ConvNet in our framework is trained to distinguish such properties and predict the transparency values, which are further combined with WSSV to generate MSI.
\begin{table}[!t]
\begin{center}
\resizebox{0.45\textwidth}{!}{
\begin{tabular}{cccccccc}
\toprule
\textbf{Layer} & \textbf{s} & \textbf{d} & \textbf{n} & \textbf{depth} & \textbf{in} & \textbf{out} & \textbf{input} \\\hline
conv1\_1 & 1 & 1 & 8  & 32/32 & 1 & 1 & WSSV           \\
conv1\_2 & 2 & 1 & 16 & 32/16 & 1 & 2 & conv1\_1         \\
conv2\_1 & 1 & 1 & 16 & 16/16 & 2 & 2 & conv1\_2         \\
conv2\_2 & 2 & 1 & 32 & 16/8  & 2 & 4 & conv2\_1         \\
conv3\_1 & 1 & 1 & 32 & 8/8   & 4 & 4 & conv2\_2         \\
conv3\_2 & 1 & 1 & 32 & 8/8   & 4 & 4 & conv3\_1         \\
conv3\_3 & 2 & 1 & 64 & 8/4   & 4 & 8 & conv3\_2         \\
conv4\_1 & 1 & 2 & 64 & 4/4   & 8 & 8 & conv3\_3         \\
conv4\_2 & 1 & 2 & 64 & 4/4   & 8 & 8 & conv4\_1         \\
conv4\_3 & 1 & 2 & 64 & 4/4   & 8 & 8 & conv4\_2         \\\hline
nnup\_5  &   &   &    & 4/8   & 8 & 4 & conv3\_3+conv4\_3 \\
conv5\_1 & 1 & 1 & 32 & 8/8   & 4 & 4 & nnup\_5          \\
conv5\_2 & 1 & 1 & 32 & 8/8   & 4 & 4 & conv5\_1         \\  
conv5\_3 & 1 & 1 & 32 & 8/8   & 4 & 4 & conv5\_2         \\  
nnup\_6  &   &   &    & 8/16  & 4 & 2 & conv2\_2+conv5\_3 \\
conv6\_1 & 1 & 1 & 16 & 16/16 & 2 & 2 & nnup\_6          \\
conv6\_2 & 1 & 1 & 16 & 16/16 & 2 & 2 & conv6\_1         \\  
nnup\_7  &   &   &    & 16/32 & 2 & 1 & conv1\_2+conv6\_2 \\
conv7\_1 & 1 & 1 & 8  & 32/32 & 1 & 1 & nnup\_7          \\
conv7\_2 & 1 & 1 & 8  & 32/32 & 1 & 1 & conv7\_1         \\  
conv7\_3 & 1 & 1 & 1  & 32/32 & 1 & 1 & conv7\_2         \\
\bottomrule
\end{tabular}
}
\end{center}
\caption{Our proposed network architecture. The size of all convolution kernels are set to 3. Column \textbf{s},\textbf{d},\textbf{n} denote the stride, kernel dilation and number of output filters respectively. \textbf{depth} column shows the number of layer input/output depth (we use 32 input depth layers as an example), while \textbf{in} and \textbf{out} are the accumulated stride of layer input/output. \textbf{input} names with + mean concatenation along the depth dimension and \textbf{Layer} names starting with “nnup” perform 2x nearest neighbor upsampling.}
\label{tab:network}
\end{table}
\vspace{-9pt}
\subsection{Network Architecture}
\label{sec:network}
Our 3D ConvNet architecture is a slight variation of Middle~\etal 's work~\cite{mildenhall2019local}. 
In our design, we modify the network to suit the WSSV input described in Sec.~\ref{sec:wssv}. 

The output vector of the neural network has a shape of [H, W, D, 1], which is converted into the $\alpha$ channel of an MSI by a ReLu function. This $\alpha$ channel is then stacked with WSSV RGB 
volume to form the MSI. As color information of an MSI comes directly from WSSV, which is the combination of projected input footage, during training, the network is 
not learning to simulate the correct color but to select the correct layer by inferring the corresponding weight of the $\alpha$ channel to imply an object's distance.

The training objective is :
\begin{equation}
\begin{split}
&{MSI}_{\theta} = {f}_{\theta}(WSSV) \\
&{\cal{L}} = {\cal{L}}_{L1} + {\lambda}_{VGG}{\cal{L}}_{VGG}\\
&{\argmin}_{\theta}~\sum_{i=1}^{M}{\cal{L}}(Render({MSI}_{\theta},\omega_c, \omega_i), {\cal{W}}_{ERP}(I_i)),
\end{split}
\label{equ:train}
\end{equation} 
where the goal is to minimize the difference between the 
rendered MSI result and ground truth. In Equ.~\ref{equ:train} the 3D ConvNet $f_\theta(\cdot)$ takes in $WSSV$ and predicts ${MSI}_{\theta}$. 
The $Render(\cdot)$ function follows the differentiable 
rendering scheme in Equ.~\ref{equ:alpha rendering}, where the ${\cal{W}}_{ERP}(\cdot)$ is the fisheye to ERP warp function. The training loss $ \cal{L}(\cdot)$ is a weighted combination of 
L1 loss ${\cal{L}}_{L1}(\cdot)$ and VGG loss $ {\cal{L}}_{VGG}(\cdot)$ proposed in~\cite{chen2017photographic}. We refer readers to Tab.~\ref{tab:network} for detailed network architecture.

\begin{table}[!t]
\begin{center}

\begin{tabular}{ccc|c|c}
\toprule
\textbf{N} & \textbf{Dataset}       & \textbf{Resolution} & \textbf{PSNR} & \textbf{SSIM} \\ \hline
\multirow{2}{*}{N=32} & \multirow{4}{*}{UnrealCV} & $400\times200$&28.28$\pm$2.45&0.90$\pm$0.03      \\
                      &                           & $640\times320$&28.79$\pm$2.40&0.90$\pm$0.03      \\
\multirow{2}{*}{N=64} &                           & $400\times200$&28.52$\pm$2.51&0.90$\pm$0.03      \\
                      &                           & $640\times320$&29.23$\pm$2.43&0.91$\pm$0.03 \\\hline
\multirow{2}{*}{N=32} & \multirow{4}{*}{Replica}  & $400\times200$&33.48$\pm$3.72&0.95$\pm$0.03       \\
                      &                           & $640\times320$&33.40$\pm$3.87&0.95$\pm$0.03      \\
\multirow{2}{*}{N=64} &                           & $400\times200$&33.59$\pm$3.74&0.95$\pm$0.03      \\
                      &                           & $640\times320$&33.80$\pm$3.92&0.95$\pm$0.03      \\ \bottomrule
\end{tabular}
\caption{Evaluation results. We examine our framework on the combination of two different number of MSI layers (\textbf{N}=32 and \textbf{N}=64), two input resolution (400$\times$200 and 640$\times$320) and two data generation engine (UnrealCV and Replica). }
\label{tab:results}
\end{center}
\end{table}
\vspace{-9pt}
\section{Experiments}
\label{sec:experiments}
We trained and evaluated our network performance by generating MSI representations and rendering them in different input positions and computing 
peak signal-to-noise ratio (PSNR) and structural similarity index (SSIM) quality scores against ground truth generated using our proposed approach in Sec~\ref{sec:data}. 
In this section, we describe network implementation details, and experiment results.
\subsection{Implementation Details}
\label{sec:implementation}
We implement our proposed 3D ConvNet using TensorFlow API following the description in Tab.~\ref{tab:network}. During training, 
Automatic Mixed Precision feature was applied to utilize GPU memory by using half precision (FP16) in computation. The network was optimized with an SGD optimizer 
with the learning rate set to $2\times{10}^{-4}$. We employed the VGG loss~\cite{chen2017photographic} as the perceptual loss with weight $\lambda_{VGG}=200$. 
Using the data generation method described in Sec.~\ref{sec:data}, we synthesized a set of 6-sensors VR camera footage on 2000 locations 
and masked the field of view to ${190}^{\circ}$ and ${220}^{\circ}$ randomly. The entire dataset was split into 1600 locations for training subsets and the reset for evaluation. 
The network was trained on 640$ \times$320 resolution with 32 layers of MSI and 400$\times$200 resolution with 64 layers of MSI simultaneously for 400k iterations on an Nvidia RTX 2080Ti GPU.
\begin{figure}[tb]
 \centering 
 \includegraphics[width=0.42\linewidth]{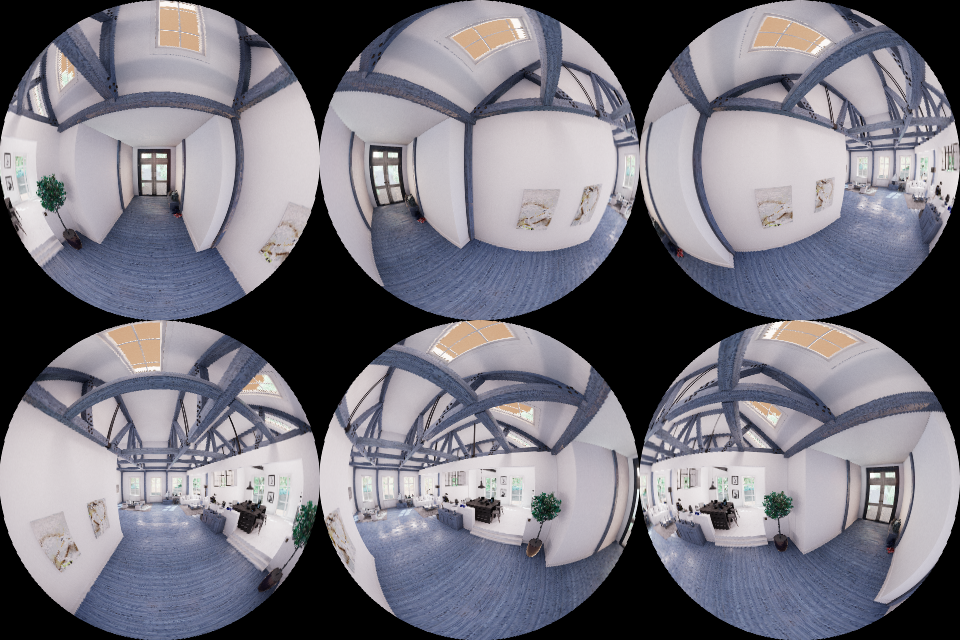}
  \includegraphics[width=0.56\linewidth]{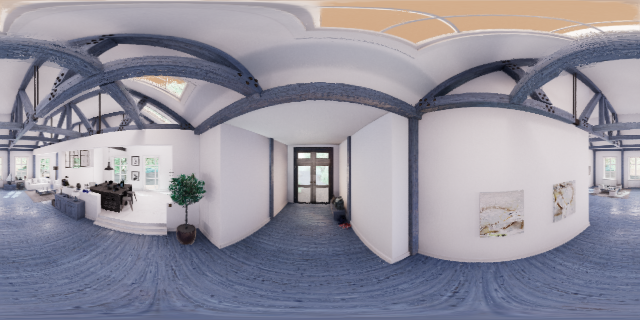}\\
   \includegraphics[width=0.42\linewidth]{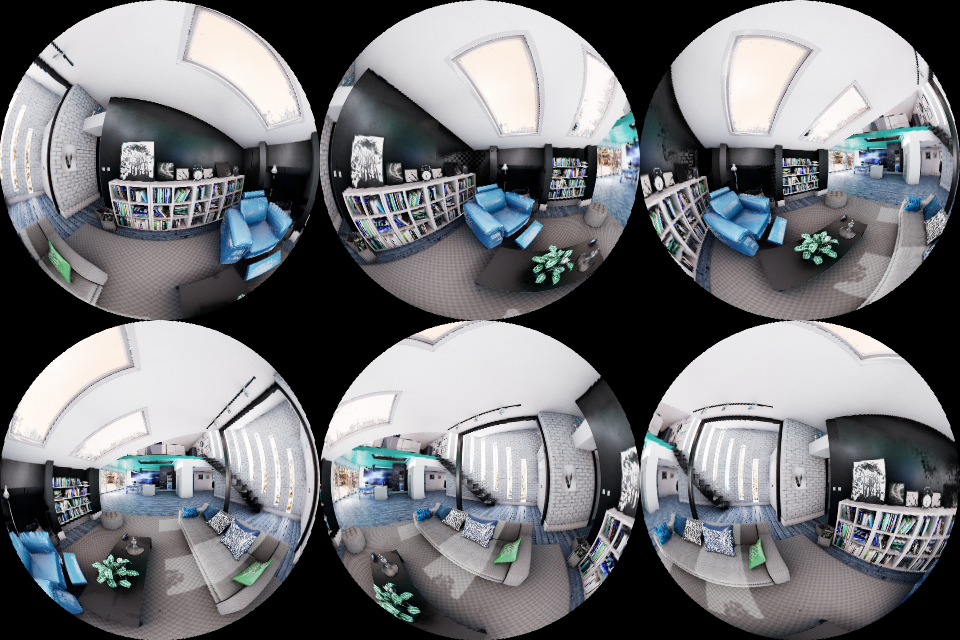}
  \includegraphics[width=0.56\linewidth]{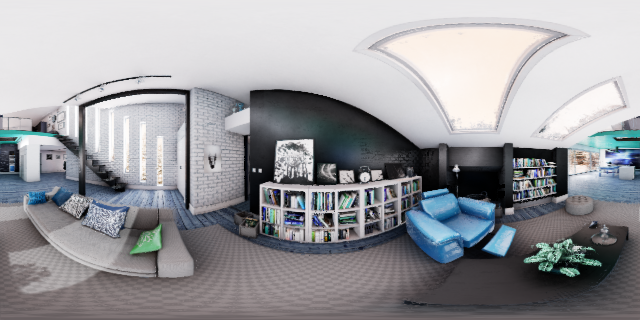}\\
   \includegraphics[width=0.42\linewidth]{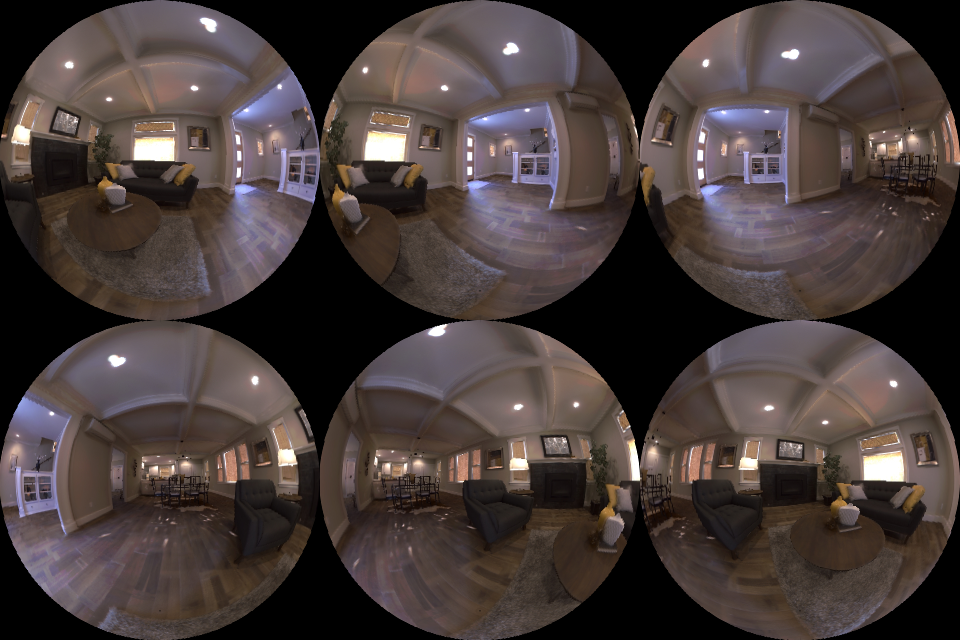}
  \includegraphics[width=0.56\linewidth]{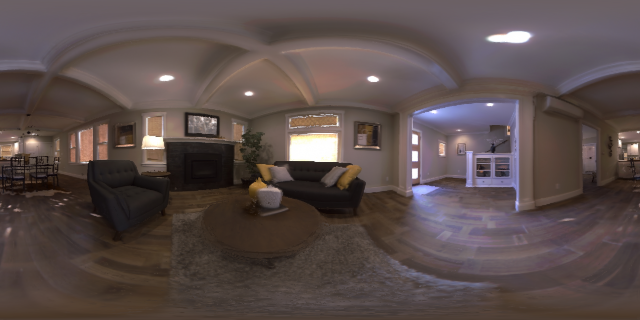}\\

 \caption{Illustration of system input and output. Left column: Input fisheye imags. Right column: Render result from MSI predicted by our system.}
 \label{fig:results}
\end{figure}
\subsection{Evaluation}
\label{sec:evaluation}
We examined the performance of our approach with the evaluation subsets that contains 400 locations with two different resolutions and corresponding 
color ground truth. We generated MSIs on evaluation subsets using our proposed method and rendered each MSI to novel viewpoints using input poses, then 
computed PSNR and SSIM of these novel views with the ground truth color. 
These quantitative results are reported in Tab.~\ref{tab:results} and some selected qualitative results are shown in Fig~\ref{fig:results}.

As demonstrated in Tab.~\ref{tab:results}, our system can generate high-quality 6-DoF contents from VR camera footage. Comparing with the ground truth color, 
results rendered on input views achieve an average PSNR over 31 dB. After carefully examining the quantitative results in Tab.~\ref{tab:results}, 
we notice a minor performance difference between two datasets. A small variance between two render engines could be the cause, as UnrealCV engine renders reflectance 
that varies from different angles while Replica uses static reconstructed object texture acquired from real-world scenes. Overall, results rendered 
from MSIs show plausible visual details on complex textures like leaves
, books, floor textures as shown in Fig.~\ref{fig:results}.
We also notice there is a performance gain by increasing the number of layers in MSI. As each MSI is a set of concentric spheres, a denser set of spheres can represent 
more detailed depth variances among scene objects. 
In our experiments, we also confirmed that our network can be trained and utilized among variable image resolution and number of layers of an MSI. 
We used 640x320 with 32-layer and 400x200 with 64-layer configuration on training and evaluated the network on the combination of both resolution and numbers of layers. Experiment results show that the network is not overfitting on one particular configuration.
\vspace{-9pt}
\section{Conclusions}
\label{sec:conclusion}
In this paper, we present an end-to-end deep-learning framework to compose 6-DoF omnidirectional with multi-sphere images. We use weighted sphere sweep volume to 
unify inputs from various panoramic camera setups into one constant volume size, solving the compatibility issue in previous work. Combined with our proposed 3D ConvNet architecture, 
we can process camera footage directly and reduce the systematic artifacts introduced in the ODS stitching process. We propose a high-quality 6-DoF dataset generation 
method using UnrealCV and Facebook Replica engines for training and quantitive performance evaluations. A series of experiments were conducted to verify our system. 
Experiment results show our system can operate on variable image resolution and MSI layers, as well as producing high-quality novel views that contain correct occlusion and detailed textures.


\bibliographystyle{IEEEbib}
\bibliography{main}

\end{document}